\documentclass{article}

\usepackage{PRIMEarxiv}
\usepackage{multirow}
\usepackage[utf8]{inputenc} 
\usepackage[T1]{fontenc}    
\usepackage{hyperref}       
\usepackage{url}            
\usepackage{booktabs}       
\usepackage{amsfonts}       
\usepackage{nicefrac}       
\usepackage{microtype}      
\usepackage{lipsum}
\usepackage{fancyhdr}       
\usepackage{graphicx}
\usepackage{amsmath}
\graphicspath{{media/}}     

\title{360-Degree Full-view Image Segmentation by Spherical Convolution compatible with Large-scale Planar Pre-trained Models}

\author{Jingguo Liu$^{1}$, Han Yu$^{2}$, Shigang Li$^{2}$, Jianfeng Li$^{1*}$
	\\
	$^{1}$Southwest University, Chongqing, China\\
	$^{2}$Hiroshima City University, Hiroshima, Japan\\
	{\tt\small popqlee@swu.edu.cn}\\
	{\tt\small $^{*}$Corresponding author}\\
}

\begin{document}
	\maketitle
	\begin{abstract}
		Due to the current lack of large-scale datasets at the million-scale level, tasks involving panoramic images predominantly rely on existing two-dimensional pre-trained image benchmark models as backbone networks. However, these networks are not equipped to recognize the distortions and discontinuities inherent in panoramic images, which adversely affects their performance in such tasks. In this paper, we introduce a novel spherical sampling method for panoramic images that enables the direct utilization of existing pre-trained models developed for two-dimensional images. Our method employs spherical discrete sampling based on the weights of the pre-trained models, effectively mitigating distortions while achieving favorable initial training values. Additionally, we apply the proposed sampling method to panoramic image segmentation, utilizing features obtained from the spherical model as masks for specific channel attentions, which yields commendable results on commonly used indoor datasets, Stanford2D3D.
	\end{abstract}
		%
	\section{Introduction}
	\label{sec:intro}
	Distortion will arise at the north and south poles in panoramic images.
	To address this issue, existing methods commonly incorporate additional modules \cite{bending-reality} or alternative image representations \cite{13,14,16} for separate processing. While these approaches have proven effective, they typically incur additional network costs and their results are constrained by the employed fusion mechanisms. Significantly, compared to traditional two-dimensional perspective images, panoramic images entail higher capture costs and substantial storage requirements, which has contributed to the scarcity of million-scale panoramic datasets suitable for large-scale model pre-training. Previous studies, such as ResNet \cite{He_2016_CVPR}, Vision Transformer \cite{dosovitskiy2020image}, and ConvNext \cite{liu2022convnet}, have demonstrated that pre-trained models developed on million-scale datasets, followed by comprehensive fine-tuning for specific tasks, exhibit superior fitting capabilities relative to models retrained from scratch. To address this gap, our objective is to leverage the weights of pre-trained models developed on existing two-dimensional perspective images and directly apply appropriate de-distortion methods to the network architecture in order to eliminate image distortion, without incurring additional branching overhead. Building on this premise, we propose an innovative approach that involves directly modifying the planar convolution operation of existing pre-trained models by artificially adjusting and correcting the sampling points of the convolution kernels to achieve de-distortion convolution.
	
	Existing methods\cite{hohonet,bending-reality,complementary-bi-directional,look-at-the-Neighbor,Yu_2023_CVPR} to eliminate panoramic distortion are mainly divided into vector-based method, learnable convolution method, different representation fusion method, and spherical convolution method. Among them, the vector-based method splits the panoramic image into a series of one-dimensional vectors, and the vectors in the same row share the same degree of distortion. However, these method weakens the local perception ability of the image. The learnable convolution method relies on the neural network to learn the sampling positions of the convolution kernel, but this method depends on the learning ability of the network, and the network itself has certain losses, which may not be perfect. The method of fusing different representations to eliminate panoramic distortion, such as fusing the cubemap representation and equirectangular projection\cite{13,14}, fusing the tangent image and equirectangular projection\cite{16}, can also handle panoramic distortion well. However, the cubemap is limited by the number of views, and each tangent image only has a small central part with 0 error, so more views need to be fused. The result will be limited by the expressive power of multi-view fusion. In contrast, the spherical convolution method is more in line with the nature of panoramic images, and spherical convolution can solve the image boundary problem well.
	Therefore, the spherical convolution method can better express panoramic images. Based on it, we propose to modify the sampling points of the convolution kernel on the pre-trained model and modify them according to the spherical rules to achieve spherical sampling on the two-dimensional plane.
	
	Additionally, to achieve more accurate semantic segmentation, we introduce a spherical branch that generates independent heatmaps for each semantic channel. These heatmaps focus on training the attention weights for each semantic channel. Considering the layout of different scenes in indoor panoramic images, where the image regions and features of various semantics are unified, we assign separate semantic channels for the image. Treating this as a one-channel classification task. The main contributions of this paper are as follows:
	
	1. We propose a spherical convolution and downsampling construction method that is compatible with large-scale planar pre-trained models. 
	In contrast to traditional methods, this approach enables us to achieve faster distortion correction and downsampling without incurring additional overhead for distortion elimination.
	
	2. For panoramic image semantic segmentation, we generate independent heatmaps for each semantic channel in the spherical branch. These heatmaps focus on training the attention weights for each semantic channel. By introducing an attention mechanism, the network can pay more attention to image regions that are relevant to specific semantics, thereby improving the accuracy and robustness of semantic segmentation.
	
	3. The experimental results demonstrate that our method achieves superior performance compared to existing approaches on the Stanford2D3D.
	\section{Related Work}
	\subsection{Spherical Convolution}
	Su et al.\cite{26} utilize transformers to transfer the convolutional kernels adopted on perspective images point-wise onto panoramic ERP format images. Perraudin et al.\cite{30} presented a spherical CNN that constructed by representing the sphere as a graph, and utilized the graph-based representation to define the standard CNN operations. These methods have provided evidence for the effectiveness of spherical convolutions in processing information from panoramic image. Yu et al.\cite{OSRT} design a distortion-aware Transformer to modulate ERP distortions continuously and self-adaptively. Lee et al.\cite{spherePHD} utilizes a spherical polyhedron to represent omni-directional views to minimizes the variance of the spatial resolving power on the sphere surface. Li et al.\cite{li2023spherical} proposes a spherical convolution-empowered FoV prediction method, which is a multi-source prediction framework combining salient features extracted from 360$^{\circ}$ video with limited FoV feedback information. Liu et al.\cite{Liu_2024_CVPR} propose a novel circular spherical convolution method and apply it in the field of panoramic depth estimation, achieving promising results.
	
	\subsection{Semantic segmentation}
	\begin{figure}[t]
		\centering
		\includegraphics[width=\linewidth]{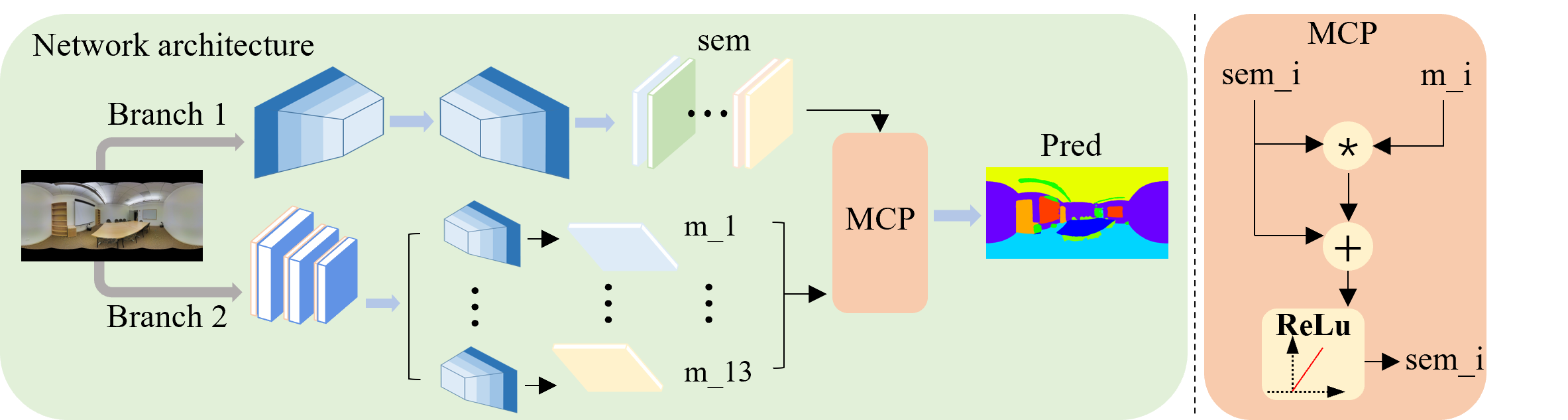}
		\caption{Network architecture}
		\label{fig:1}
	\end{figure}
	Recently, \cite{Orientation-aware} and \cite{tangent-images} used different projection formats to mitigate distortion. Tateno et al.\cite{tateno2018distortion} propose a learning approach for panoramic depth map estimation from a single image and demonstrate their approach for emerging tasks such as panoramic semantic segmentation. Yang et al.\cite{yang2019pass} propose a Panoramic Annular Semantic Segmentation framework to perceive the whole surrounding based on a compact panoramic annular lens system and an online panorama unfolding process. 
	Sun et al.\cite{hohonet} utilized horizontal features to obtain a comprehensive understanding of indoor 360-degree panoramas. Yu et al.\cite{Yu_2023_CVPR} present PanelNet, a framework that understands indoor environments using a novel panel representation of 360 images. Zhang et al.\cite{bending-reality} propose to learn object deformations and panoramic image distortions in the Deformable Patch Embedding and Deformable MLP components which blend into Transformer for PAnoramic Semantic Segmentation model. Zheng et al.\cite{complementary-bi-directional} combine the two different representations and propose a novel 360-degree semantic segmentation solution from a complementary perspective. Zheng et al.\cite{both-style-and-distortion-matter} propose a novel yet flexible dual-path UDA framework, DPPASS, taking ERP and tangent projection images as inputs. 
	
	\section{Method}
	\subsection{Overview}
	As Figure 1 shows, we proposes a novel spherical convolutional pre-training model and downsampling method that is compatible with existing large-scale planar pre-trained models. This approach enables direct utilization of mature pre-trained models trained on planar datasets, allowing for ultra-rapid graph adaptation fine-tuning with minimal training overhead, thereby enhancing the effectiveness of network study. Additionally, since the down-sampling method designed based on this pre-training model cannot be directly applied to the model's up-sampling stage, we have devised a novel decoder module specifically for panoramic semantic segmentation tasks. This decoder splits the spherical branch into individual semantic channels to generate separate attention maps, denoted as $m_i$. These attention maps focus on training attention weights for each semantic channel, similar to expert models that specialize in detecting and segmenting their respective specific classes. This facilitates more accurate recognition of each class within the network. Specifically, for the network input, we first perform coarse semantic segmentation, denoted as $sem$, using the original planar pre-trained network. Simultaneously, the same input is fed into our designed panoramic image pre-training network, which can directly leverage the weights of the planar pre-trained model. Through a simple up-sampling network, we predict masks for each channel. Finally, we adopt a processing method similar to an attention mechanism, where the obtained masks and $sem$ are calculated on a per-channel basis to get the final segmentation results.Similar to previous panoramic semantic segmentation tasks\cite{complementary-bi-directional}, we adopt cross-entropy loss to regularize our results.
	
	\subsection{Pretrained spherical model}
	Different with conventional perspective images, pre-trained models based on panoramic images are not readily available.
	Therefore, additional distortion elimination methods, such as fusion\cite{13,14,16}, deformable convolutions\cite{bending-reality,behind-every-domain}, and spherical convolutions\cite{Liu_2024_CVPR}, are often required to correct distortions in panoramic feature maps. Pre-trained models, however, offer more reliable initial values for network parameters, facilitating rapid convergence to optimal solutions. This is because the network has already gained a profound understanding of images and their features through extensive training on large datasets, accelerating its ability to adapt to new tasks. Inspired by previous works\cite{Liu_2024_CVPR}, spherical convolutions align better with the nature of spherical surfaces. If we can adapt large planar pre-trained models directly using spherical convolutions, treating the network as a spherical processor, we can achieve better results. Specifically, in this paper, we adopt ConvNext as the pre-trained model, which utilizes convolutional kernels of sizes 2, 4, and 7. For $2\times2$ and $4\times4$ convolutions, it employ strides equal to the kernel size, while the $7\times7$ convolution has a stride of 1.
	
	To adapt these convolution kernels to panoramic images, we manually modify the image. Taking the $4\times4$ convolution kernel as an example, in the planar domain, it divides the image into $4\times4$ blocks for convolution. For panoramic images, we project the panoramic image onto a spherical surface and select 16 points, denoted as $p_1 \cdots p_{16}$, on the sphere to form a new convolution kernel. To utilize the pre-trained ConvNext, we reproject the image so that the sampling points used by the planar convolution kernels directly correspond to pixels sampled on the spherical surface, rather than traditional 2D planar points. The process effectively identifies the true neighboring positions for the center point, eliminating spherical distortion. (The $2\times2$ and $4\times4$ convolution kernels are handled similarly.)
	As illustrated in Figure 2, for a convolution operation using a $4\times4$ kernel with a stride of 4, we assume the image can be partitioned into six blocks, each blocks is formed by a $4\times4$ convolutional kernel. Subsequently, we can rearrange the pixels within each block using the pixel values from the original image. 
	\begin{equation}
		\begin{aligned}
			u^\prime &=\frac{W}{2\pi}arctan2\left(y^\prime,x^\prime\right) \qquad 
			v^\prime &=\frac{H}{\pi}arccos{(z}^\prime)
			\label{e_2}
		\end{aligned}
	\end{equation}
	Two circles are then selected on the sphere to form the $4\times4$ kernel. Similar to Liu et al.\cite{Liu_2024_CVPR}, we initially consider the north pole as the center, but instead of directly selecting the spherical kernel at the north pole as shown in Figure 2, we utilize the 2.5th row and column as a center, which is not part of the convolution kernel itself. Consequently, we draw two circles centered at the North Pole, with the inner circle having a radius of $\frac{2\pi}{\frac{W}{2}}$, Where $W$ denote the width of the ERP image, and spacing between its four points of $\frac{\pi}{4}$, and the outer circle having a radius of $\frac{2\pi}{\frac{3W}{2}}$ with spacing between its twelve points of $\frac{\pi}{12}$. Next, we rotate all points of the kernel along with the north pole to the spherical coordinates($x^\prime,y^\prime,z^\prime$) corresponding to our chosen center. This process yields the spherical coordinates for the respective convolution kernel. Subsequently, we project all points of the kernel to the ERP format using Equation \ref{e_2} to obtain their coordinates in the kernel's position. For non-integer coordinates, we employ bilinear interpolation to determine their values. This enables us to determine the coordinates of each selected spherical convolution kernel in the ERP image, which are then arranged into the original 4x4 matrix.
	
	\begin{figure}[t]
		\centering
		\includegraphics[width=0.8\linewidth]{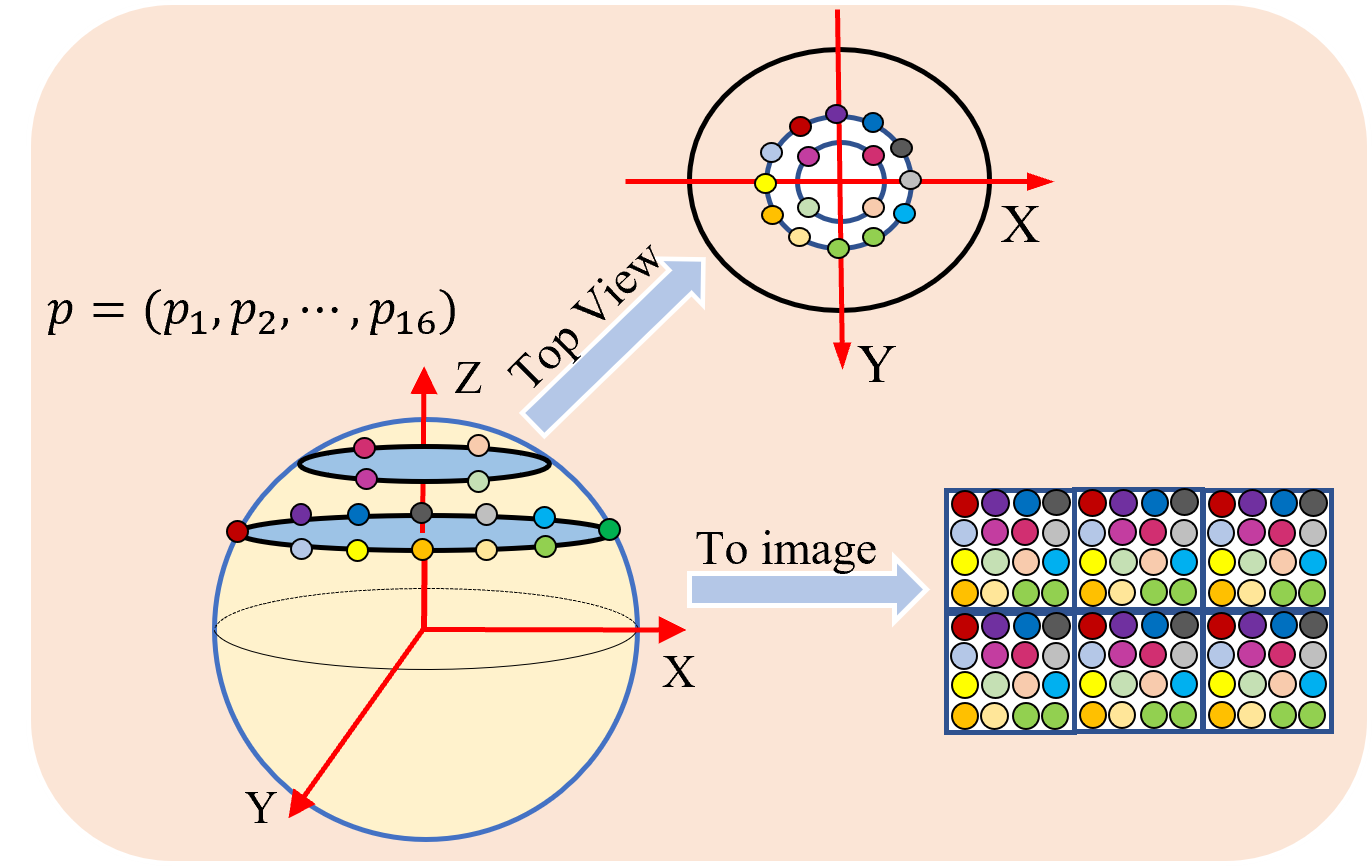}
		\caption{Spherical sampling}
		\label{fig:1}
	\end{figure}
	For $7\times7$ convolution, the kernel size is 7 with a stride of 1. Inspired by the $4\times4$ convolution kernel, we choose to construct an image that is 7 times larger than the original without overlap. Specifically, similar to the previous $4\times4$ construction method, we calculate the positions of adjacent $7\times7$ convolution kernels for each pixel location on the spherical model. Given the odd-sized convolution kernel, the North Pole serves as one of the points in the kernel. Outside the North Pole, three circles are selected with radii of $(\frac{2\pi}{W}$, $\frac{4\pi}{W}$, $\frac{6\pi}{W}$, Where $W$ denote the width of the ERP image), and inter-point distances of ($\frac{\pi}{8}$), ($\frac{\pi}{16}$), ($\frac{\pi}{24}$), respectively. Then, the selected convolution kernel points on the sphere are back-projected onto the ERP image using Equation (1). Notably, this operation implies that each pixel in the original 2D image is expanded into 49 pixels, representing the original $7\times7$ convolution kernel. At this point, we modify the sliding stride of the original $7\times7$ convolution to 7, enabling the convolution operation to seamlessly utilize ConvNeXt's original pre-trained weights. Furthermore, the process of computing the positions of image projection points can be pre-computed offline and stored as a lookup table. As described above, by artificially modifying the arrangement of image pixels to form a new image that caters to the original pre-trained model, our method, although leveraging pre-trained weights trained on ordinary images, enables sampling on the spherical surface to achieve the goal of eliminating panoramic distortion.
	
	By artificially modifying the pixel arrangement of the image to form a new image that aligns with the original pre-trained model, our method utilizes pre-trained weights while sampling on the spherical surface to eliminate panoramic distortions.
	
	\subsection{Semantic branch}
	We employ the aforementioned pre-training model, inherently tailored for panoramic image, to extract image features that effectively mitigate panoramic distortions. These features are systematically acquired through sampling on a spherical surface, leveraging circular spherical convolution kernels. In the selection of these kernels, we meticulously determine their radii, grounded in the image width, to ensure precise representation of the distances between neighboring pixels along any meridian within the spherical domain. Despite the utilization of decimal precision, discrete spherical computations inherently give rise to computational rounding errors, futhermore, at the deep networks, the top line of the image can not represent the North Pole. These errors exhibit a tendency to intensify in deeper network architectures during iterative convolution processes, exacerbated by geometric distortions where a regular spherical shape manifests as synaptic patterns at the image boundaries, resulting in the dispersion of convolutional feature maps towards the poles in the 2D projection.
	
	Since the spherical convolution present a challenges, particularly in implementing up-sampling, we propose a dual-branch architecture. As depicted in Figure 1, our approach incorporates the aforementioned spherical-compatible pre-training model as the secondary branch, juxtaposed with a fully planar pre-training model constituting the primary branch. Consequently, the backbone of our network maintains the precision of the planar pre-training model, functioning as a robust reference for image perception and prediction. Furthermore, drawing inspiration from attention mechanisms, we introduce a spherical branch that theoretically aligns more closely with spherical properties, serving as a feature attention weight. This branch acts as an attention enhancer, prioritizing the network's ability to comprehend contextual information over precise contour details. By integrating this spherical attention mechanism, our network becomes more adept at capturing the intricacies of panoramic imagery, thereby refining predictions and enhancing overall performance.
	\begin{equation}
		Sem=Relu\left(sem_i+sem_i\ast m_{i}\right)
		\label{eq_3}
	\end{equation}
	Specifically, with respect to the features derived from the spherical module, we systematically upsample these features in a layer-wise fashion and subsequently predict a channel $x$ for each upsampled feature set. These predictions are then compared with the predictions S generated by the backbone branch through a per-channel attention mechanism. As shown in Equation \ref{eq_3}, we perform a dot product operation between the predictions obtained from each individual channel and the overall predictions, thereby emphasizing the significance of results pertaining to the specific target channel predicted by each single-channel decoder. Following this, we incorporate these emphasized results directly into the original prediction outcomes, effectively augmenting the weight assigned to the corresponding channels. To ensure the integrity of the results, we employ an additive-based summation algorithm and apply the ReLU property to eliminate any negative component outcomes, ensuring that only positive contributions are aggregated and utilized in the final prediction.
	
	
	


	\section{Experiments}
	\subsection{Datasets and Implementation Details}
	
	{\bf Stanford2D-3D dataset:} \cite{joint-2d-3d-semantic-data} consists of 1,413 equirectangular RGB images captured from six large-scale indoor areas. 
	
	\textbf{Implementation details:} Due to hardware limitations, we were unable to complete full model training at the $512\times1024$ resolution. Therefore, we conducted comparative training using an incomplete model on $512\times1024$ resolution and make the experiments on $256\times512$ resolution.
	\begin{figure}[h]
		\centering
		\includegraphics[width=0.9\linewidth]{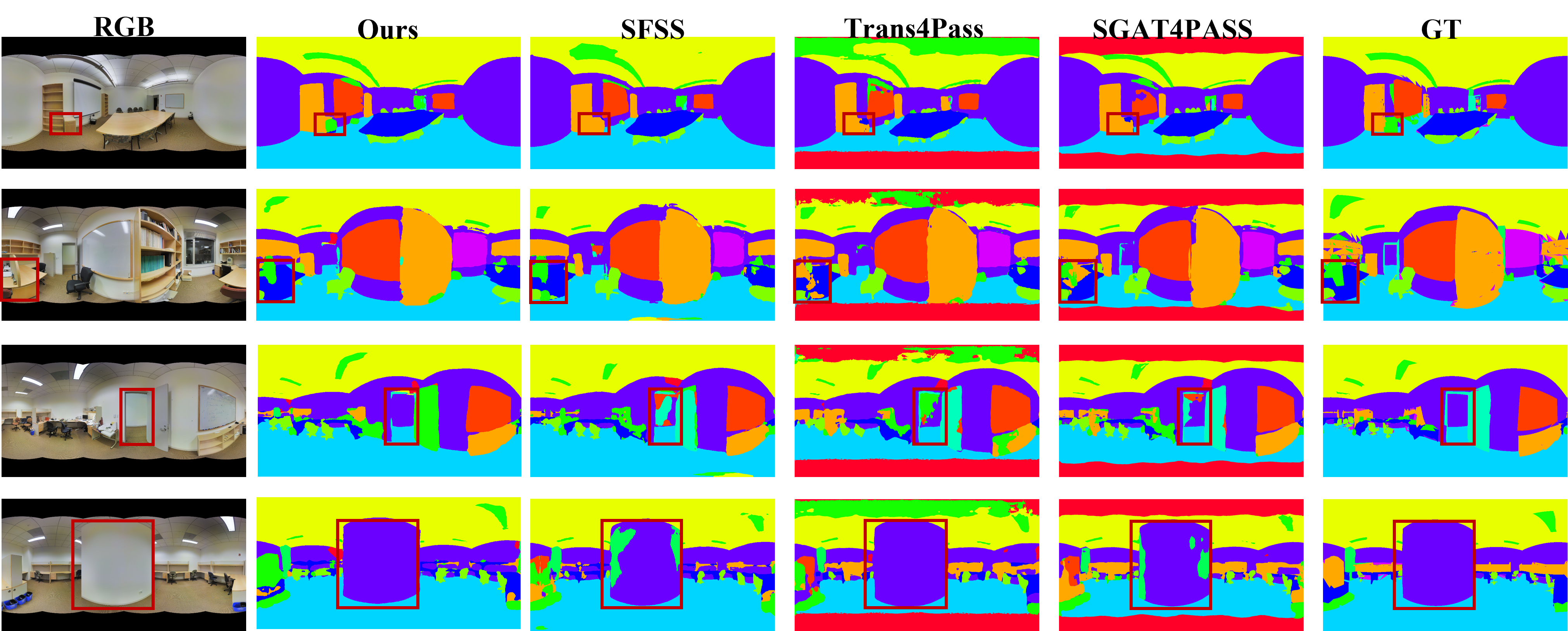}
		\caption{The segmentation results on the Stanford2D3D\cite{joint-2d-3d-semantic-data}.}
		\label{fig:2}
	\end{figure}
	\begin{figure}[t]
		\centering
		\includegraphics[width=0.9\linewidth]{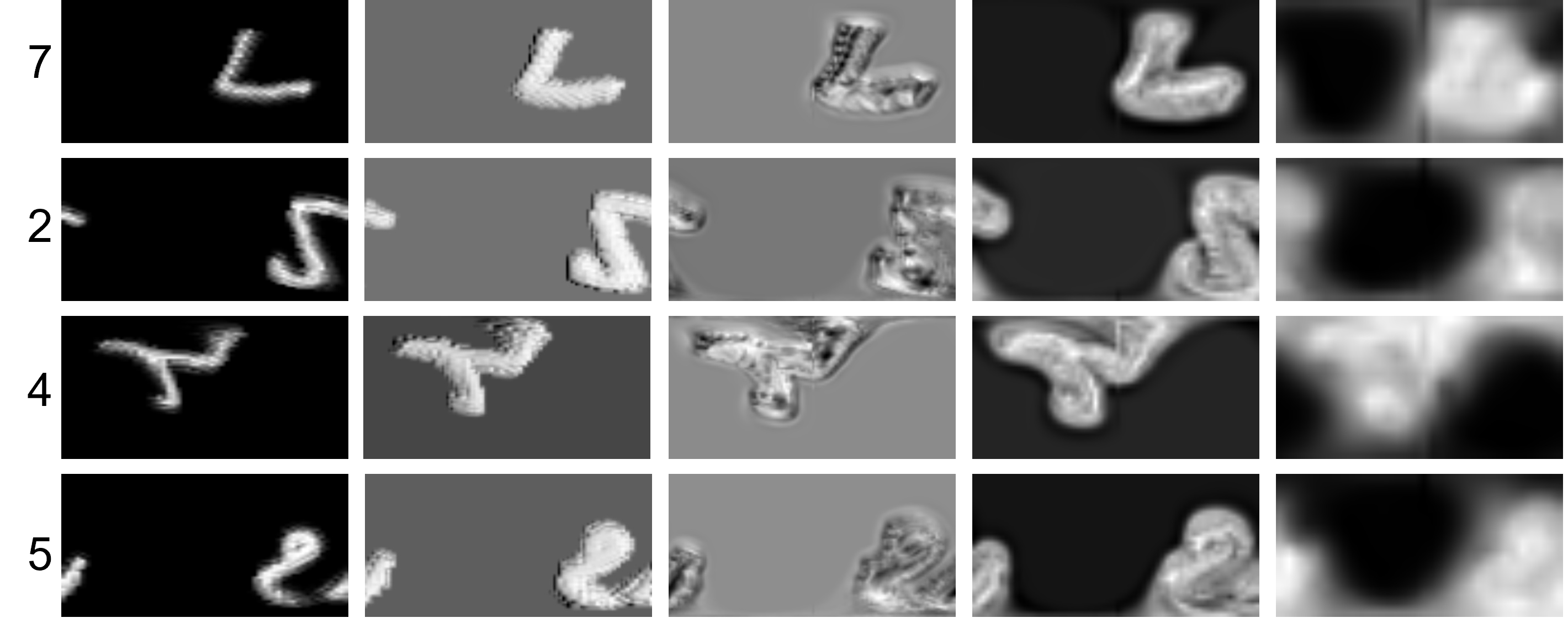}
		\caption{visualize of pre-trained model convolutions. The results presented from left to right represent the consecutively down-convolved features.}
		\label{fig:1}
	\end{figure}
	
	\subsection{Experiment Results and Analysis}
	\begin{table}[t]
		\caption{Quantitative evaluation on Stanford2D3D dataset.}
		\label{table 1} 
		\centering
		\scalebox{0.9}{%
			\begin{tabular}{c c c|c c}
				\hline
				H $\times$ W & Input & Method & mIoU$\uparrow$ & mAcc$\uparrow$ \\
				\hline
				\multirow{5}{*}{256 $\times$ 512} & RGB-D & TangentImg \cite{tangent-images} & 41.8 & 54.9 \\
				& RGB-D & HoHoNet \cite{hohonet} & 43.3 & 53.9 \\
				& RGB-D & PanoFormer \cite{PanoFormer} & 48.9 & 64.5 \\
				& RGB-D & CBFC \cite{complementary-bi-directional} & 53.8 & 66.5 \\
				& RGB-D & Ours & \bf{53.91} & \bf{67.22} \\
				\hline
				\multirow{2}{*}{256 $\times$ 512} & RGB & PanelNet \cite{Yu_2023_CVPR} & 46.3 & 58.7 \\
				& RGB & Ours & \bf{52.03} & \bf{64.02} \\
				\hline
				\multirow{9}{*}{512 $\times$ 1024} & RGB & TangentImg \cite{tangent-images} & 45.6 & 65.2 \\
				& RGB & HoHoNet \cite{hohonet} & 52.0 & 63.0 \\
				& RGB & PanoFormer \cite{PanoFormer} & 52.4 & 64.3 \\
				& RGB & CBFC \cite{complementary-bi-directional} & 52.2 & 65.6 \\
				& RGB & SFSS \cite{Single-frame-semantic-segmentation} & 52.87 & 63.96 \\
				& RGB & Trans4PASS \cite{bending-reality} & 52.10 & $-$ \\
				& RGB & Trans4PASS+ \cite{behind-every-domain} & 53.70 & $-$ \\
				& RGB & 360BEV \cite{teng2024360bev} & 54.30 & $-$ \\
				& RGB & SGAT4PASS \cite{sgat4pass} & \bf{55.30} & $-$ \\
				& RGB & Ours & 54.58 & \bf{66.33} \\
				\hline
			\end{tabular}%
		}
	\end{table}
	We perform empirical evaluations on Stanford2D3D\cite{joint-2d-3d-semantic-data}. It is worth noting that, due to multiple attempts during the model replication process, we were unable to reproduce the model performance as described in the SGAT4PASS\cite{sgat4pass}. Therefore, in the presentation of result figures, we have displayed our own replicated results. However, Table \ref{table 1} have used the original results from the SGAT4PASS. As shown in Table \ref{table 1}, we compared our method with several existing approaches. At the $256\times512$ resolution,
	with RGB images as input, our method outperformed PanelNet\cite{Yu_2023_CVPR} by 5.73 in mIoU and 5.32 in mAcc. When using RGB-D format images, our method showed significant improvements compared to traditional approaches like tangentimg\cite{tangent-images}, HoHoNet\cite{hohonet}, and PanoFormer\cite{PanoFormer}, achieving a notable 5.01 increase in mIoU compared to PanoFormer. Compared to the recent CBFC\cite{complementary-bi-directional} method, our method also showed substantial improvements, achieving approximately 0.11 improvement in mIoU and 0.72 improvement in mAcc.
	At the $512\times1024$ resolution, similar to previous methods, we only use RGB format as input. Due to hardware limitations, we are unable to employ the entire network for training in this resolution. Despite using an incomplete structure for model training at this scale, our subsequent ablation experiments demonstrated that a fully structured model has significant potential for improvement over existing results at this resolution. Specifically, compared to traditional tangentimg\cite{tangent-images} and HoHoNet\cite{hohonet}, our method showed considerable improvements. Compared to PanoFormer\cite{PanoFormer}, our results increased by 2.18 in mIoU, outperformed CBFC\cite{complementary-bi-directional} by 2.38 in mIoU, and 0.73 in mAcc. Furthermore, compared to SFSS\cite{Single-frame-semantic-segmentation}, our method outperformed by approximately 1.7 in mIoU and outperformed TransPASS+\cite{behind-every-domain} by 0.88 in mIoU. Compared to 360BEV\cite{teng2024360bev}, our method outperformed 360BEV\cite{teng2024360bev} by 0.28 in mIoU. Even when compared to SGAT4PASS\cite{sgat4pass}, our method achieved competitive results.
	\subsection{Qualitative analysis}
	We conducted a visual comparison using the results from our incomplete model training.  As depicted in Figure 3, the second column shows our results, the third column presents the results of SFSS\cite{Single-frame-semantic-segmentation}, the fourth column displays the results of Transpass\cite{bending-reality}, and the fifth column showcases the outcomes of SGAT4PASS\cite{sgat4pass} (notably, the results of Transpass and SGAT4PASS were reproduced by us). Our method achieves competitive results with existing methods.
	
	\subsection{Ablation Studies}
	\textbf{Effectiveness of spherical pre-trained model}
	The spherical representation of handwritten digit recognition can lead to indistinguishable outcomes. If sampling on the spherical model is not conducted appropriately, even such a simple classification task may fail to yield satisfactory results. To demonstrate the efficacy of the proposed spherical pre-training model, similar to previous spherical models\cite{cohen2018spherical,spherePHD}, we conducted ablation experiments on the MNIST to validate the model's effectiveness. Under the same training settings, the prediction accuracy without utilizing pre-trained model weights was 66\%, whereas incorporating pre-trained weights significantly boosted the accuracy to 96\%. The jump from 66\% to 96\% in prediction accuracy when incorporating pre-trained weights into the spherical pre-training model highlights the significant potential of this approach. Such a significant improvement highlights the effectiveness of utilizing pre-trained knowledge within the framework of spherical convolutions. It demonstrates the model's enhanced capability to adapt to and comprehend panoramic distortion images. This outcome also emphasizes the exceptional compatibility of our spherical pre-training model with existing pre-trained weights, thereby enabling superior recognition performance for panoramic distortion images through strategic spherical model sampling.
	
	\begin{table}[h]
		\caption{Ablation study on the number of network layers and fusion methods}
		\label{table3}
		\begin{center}
				\begin{tabular}{cc c c c c|c}
					\hline
					Base & layer1 & layer2 & layer3 & layer4 & Cat & mIoU$\uparrow$ \\
					\hline
					$\checkmark$ &&&&&&50.51 \\
					$\checkmark$ &$\checkmark$&&&&&50.91 \\
					$\checkmark$ &$\checkmark$&$\checkmark$&&&& 51.99\\
					$\checkmark$ &$\checkmark$&$\checkmark$&$\checkmark$&&&\textbf{52.66} \\
					$\checkmark$ &$\checkmark$&$\checkmark$&$\checkmark$&$\checkmark$&&51.31\\
					$\checkmark$ &$\checkmark$&$\checkmark$&$\checkmark$&&$\checkmark$&52.13\\
					\hline
				\end{tabular}
		\end{center}
	\end{table}
	Furthermore, to show the effectiveness of spherical discrete sampling, we visualized the features at each layer, as depicted in the Figure 4. From the visualization, it becomes evident that as the network depth increases and feature dimensions decrease, due to the discrete nature of spherical projection, image features gradually disperse towards the poles of the sphere. Specifically, as convolutional down-sampling progresses, the feature size gradually shrinks. At this point, the first row of the image can no longer represent a single point, namely the North Pole of the sphere, whereas the fixed projection mode of discrete spherical projection still treats it as a single point. This results in the phenomenon of image features diffusing towards the poles, leading to certain losses in pixel-level tasks and causing inconveniences in the end-to-end network usage of the model. Therefore, an appropriate inverse diffusion method needs to be adopted to perform inverse diffusion operations on the model during the up-sampling process.
	\textbf{Effectiveness of segmentation module}
	To determine the optimal number of spherical downsampling layers to use as a feature mask for model selection, we conducted ablation experiments. During the downsampling process of the model, we obtain 4 sets of features with varying dimensions compared to the original image size. Taking a $256\times512$ original input as an example, we obtain feature maps of sizes $64\times128$, $32\times64$, $16\times32$, and $8\times16$. We designated these feature sets as Layer 1, Layer 2, Layer 3, and Layer 4, from the highest to the lowest resolution, and conducted experiments accordingly. The experimental results, as shown in the Table \ref{table3}, indicate that using Layer 3 yields the best performance. However, when progressing to Layer 4, the results deteriorate due to the excessively small image features, which become overly discretized when projected onto the spherical surface. Consequently, we opted for Layer 3 as the optimal choice. Notably, the previously mentioned incomplete model was also based on this layer selection. Specifically, for images of size $512\times1024$, due to hardware constraints, we only used Layer 1 for model training and testing, which served as the final result for this resolution. Furthermore, we conducted ablation experiments on the fusion method of segmentation results. Specifically, we modified our Mask Attention Module to the conventionally used fusion method, Cat, in order to observe the effectiveness of this module. When the fusion method was changed to Cat, the performance of the model decreased significantly, albeit marginally. Nevertheless, this decrease serves as evidence for the validity of the proposed module.

	\section{Conclusion}
	In this paper, we propose a novel panoramic image down-sampling approach that is compatible with existing large-scale pre-trained models designed for 2D images.
	Subsequently, we validate the effectiveness of our approach on MNIST and further extend its application to panoramic semantic segmentation, with experimental results demonstrating competitive performance. 
	\bibliographystyle{unsrt}  
	\bibliography{references}

\end{document}